\documentclass[draftclasofoot, onecolumn]{IEEEtran}
\IEEEoverridecommandlockouts

\usepackage[noadjust]{cite}
\usepackage[bookmarks=true,breaklinks=true,colorlinks,linkcolor=red,citecolor=blue,urlcolor=BlueViolet]{hyperref}
\usepackage{multirow}
\usepackage{amsmath,amssymb,amsfonts}
\usepackage[ruled,vlined]{algorithm2e}
\newcommand{\mycommentstyle}[1]{\color[HTML]{0671b9}{#1}}
\SetKwComment{Comment}{\mycommentstyle{// }}{}
\usepackage{graphicx}
\usepackage{threeparttable}
\usepackage{subfigure}
\usepackage{textcomp}
\usepackage{xcolor}
\def\BibTeX{{\rm B\kern-.05em{\sc i\kern-.025em b}\kern-.08em
    T\kern-.1667em\lower.7ex\hbox{E}\kern-.125emX}}
\begin{document}

\title{High-Dimensional Sparse Data Low-rank Representation via Accelerated Asynchronous Parallel Stochastic Gradient Descent}


\author{
    Qicong Hu\textsuperscript{*} \quad Hao Wu\textsuperscript{*}\thanks{\textsuperscript{*}College of Computer and Information Science, Southwest University, Chongqing, China (cshqc98@gmail.com, haowuf@swu.edu.cn)}
}

\maketitle

\begin{abstract}
Data characterized by high dimensionality and sparsity are commonly used to describe real-world node interactions. Low-rank representation (LR) can map high-dimensional sparse (HDS) data to low-dimensional feature spaces and infer node interactions via modeling data latent associations. Unfortunately, existing optimization algorithms for LR models are computationally inefficient and slowly convergent on large-scale datasets. To address this issue, this paper proposes an \textit{Accelerated Asynchronous Parallel Stochastic Gradient Descent (A$^{2}$PSGD) for High-Dimensional Sparse Data Low-rank Representation} with three fold-ideas: a) establishing a lock-free scheduler to simultaneously respond to scheduling requests from multiple threads; b) introducing a greedy algorithm-based load balancing strategy for balancing the computational load among threads; c) incorporating Nesterov's accelerated gradient into the learning scheme to accelerate model convergence. Empirical studies show that A$^{2}$PSGD outperforms existing optimization algorithms for HDS data LR in both accuracy and training time.
\end{abstract}


\section{Introduction}
High-dimensional and sparse (HDS) data are pervasive in various real-world applications~\cite{Luo15, Zhong5, Wang7, Luo16, Wu13, ray2021various, Tugnait, Luo17, Bi18}, e.g. social network analysis~\cite{liu2023constraint, 9779537, 10012357, 9340571, 9865020}, recommendation systems~\cite{wu2022graph, chen2023bias, xin2019non, cui2020personalized, 9411671, gao2024causal}, bioinformatics~\cite{hu2021distributed, poleksic2023hyperbolic, zhang2023graph, 10164211, ai2023low, gao2023predicting} and image analysis~\cite{shen2023matrix, 10173648, 10271326, 10004837, Xie_2023_ICCV, he2023improved}. Efficiently extracting latent patterns from HDS data is crucial for understanding complex node interactions~\cite{luo2019temporal,qin2023adaptively, luo2021novel, qin2023parallel, 9659145}. However, these datasets are characterized by most of the values being missing, presenting significant challenges in terms of storage, computation, and analysis~\cite{10555245,10035508,10179251,9932678,9551506, 9783168,9839318,9894115,10113599}.

Low-rank representation (LR)~\cite{yuan2024fuzzy, chen2024generalized, xue2021spatial, agarwal2020flambe, 9590452, zeng2024novel, 10159989,qin2023asynchronous,10380219} has emerged as a powerful approach to address these challenges by mapping HDS data to low-dimensional spaces while preserving essential structural information. This technique enables the inference of hidden relationships between nodes and reduces the computational burden. In recent years, many efficient and accurate LR models have been proposed~\cite{wu2023mmlf,li2023generalized, luo2022neulft, wu2020advancing, luo2021adjusting, wu2020unified, fan2021lighter}. For instance, Luo \textit{et al.} \cite{luo2021novel} propose an ADMM-based LR model for knowledge extraction from HDS data generated by dynamically weighted directed networks. Wu \textit{et al.}~\cite{wu2020advancing} propose a non-negative tensor LR model for dynamic HDS data and improves the representation ability of this LR model through diverse regularization schemes. Chen \textit{et al.}~\cite{chen2020efficient} propose an efficient neural LR model to efficitently learn model parameters from HDS data with low time complexity. However, finding optimal low-rank representations is computationally intensive, especially for large-scale datasets. Existing optimization algorithms~\cite{shi2022dual, pilaszy2010fast, shi2022particle, chen2017efficient, gemulla2011large} often suffer from slow convergence rates and inefficiencies, limiting their applicability in real-time or large-scale scenarios.

Stochastic Gradient Descent (SGD) and its variants have been widely used for parallel and distributed optimization in LR models due to their simplicity and scalability~\cite{2783258, jin2016gpusgd,li2017msgd,9001229,chin2015fast, 9664622, 10155460, wu2023dynamic, 10580535}. For instance, Recht \textit{et al.}~\cite{recht2011hogwild} propose the Hogwild! algorithm, which completely randomly select instances in HDS data on multiple threads for parallel computation, but it suffers from serious overwriting problems when the data is dense. Gemulla \textit{et al.}~\cite{gemulla2011large} propose a distributed SGD (DSGD)-based LR model that blocks the HDS data into multiple disjoint sub-blocks and distributes them to machines to realize distributed computing. Zhuang \textit{et al.}~\cite{zhuang2013fast} propose a fast parallel SGD (FPSGD)-based LR model that adopts a blocking approach similar to DSGD, but the computational tasks of each parallel unit are scheduled through a scheduler. Luo \textit{et al.}~\cite{luo2012parallel} propose an alternating SGD (ASGD)-based LR model, which decouples the LR model parameter update process into two independent subtasks and realizes the parallel training of LR model parameters. Current parallel SGD algorithms encounter difficulties with HDS data due to explicit batch synchronization or locking mechanisms, hindering full utilization of hardware resources for accelerated training~\cite{9099403, gemulla2011large, qin2023adaptively, 9727662}. Asynchronous parallel optimization has shown promise in enhancing SGD efficiency by enabling  multiple processors to update an LR model's parameters simultaneously. However, balancing the computational load across threads and ensuring rapid convergence remain open challenges.

To address this issue, we propose an \textit{Accelerated Asynchronous Parallel Stochastic Gradient Descent (A$^{2}$PSGD) for High-Dimensional Sparse Data Low-rank Representation}. Our approach incorporates the Nesterov's accelerated gradient (NAG) technique~\cite{qu2019accelerated}, which is known for enhancing the convergence rate of gradient-based methods, with an innovative load balancing strategy designed to optimize parallel computations. By leveraging asynchronous parallel optimization, A$^{2}$PSGD achieves efficient representation learning for HDS data.

This paper makes several key contributions:

\begin{itemize}
    \item \textbf{Lock-free Scheduler Design}: Eliminating the explicit locking mechanism in existing schedulers via a novel asynchronous dynamic scheduling scheme, thus reducing the waiting time of threads for block scheduling.
    \item \textbf{Load Balancing Strategy}: A greedy algorithm-based load-balancing strategy is introduced to distribute computational tasks evenly across threads, mitigating the bucket effect and enhancing the model's convergence speed.
    \item \textbf{Accelerated Optimization Scheme}: Incorporating NAG into the learning scheme, thus further accelerating the model convergence.
\end{itemize}

Comprehensive empirical studies on large-scale datasets are conducted, demonstrating that the model significantly outperforms existing parallel LR models in terms of both accuracy and training time.

The remainder of this paper is organized as follows. Section~\ref{II} reviews problem modeling on low-rank representations and the SGD algorithm for HDS data. Section~\ref{III} presents the detailed design of the A$^{2}$PSGD-based LR model. Section~\ref{IV} describes experimental setting and results. Finally, Section~\ref{V} concludes the paper.

\section{Preliminaries}
\label{II}
\subsection{Problem Modeling}

\subsubsection*{{\bf Definition 1} -- An HDS matrix}
Given two sets of nodes $U$ and $V$, interaction between $U$ and $V$ is modeled as an matrix $R^{|U|\times{|V|}}$. $r_{uv}$ denotes that the $u\in U$-th node interacts with the $v\in V$-th node weight.

\subsubsection*{{\bf Definition 2} -- An LR model}
An LR model maps an HDS matrix into two low-rank feature matrices $M^{|U|\times{|D|}}$ and $N^{|V|\times{|D|}}$, where $D$ is a feature dimension, far less than $|U|$ or $|V|$.

For accurate representation of the low-rank features of an HDS matrix, a loss function is essential to guide an LR model's parameter optimization. The commonly used sum of squared errors loss function with $L_2$ regularization measures the difference between known and predicted values in the HDS matrix, formulated as follows~\cite{8941240, 9159907, 3472456}:

\begin{small}
\begin{equation}
    \varepsilon(M,N) = \frac{1}{2}\sum_{r_{uv}\in\Omega} \left( \left( r_{uv} - \langle m_{u},n_{v} \rangle \right)^{2} + \lambda \left( \| m_{u} \|_{2}^{2} + \| n_{v} \|_{2}^{2} \right) \right),
\end{equation}
\end{small}
where $\Omega$ is the set of known instances in an HDS matrix, $m_{u}$ is the $u$-th row vector of $M$, $n_{v}$ is the $v$-th row vector of $N$, $\left \| \cdot  \right \|_2$ is the $L_2$ norm of the model parameters, $\left \langle \cdot , \cdot \right \rangle$ is the inner product of two vectors, and $\lambda$ is the regularization coefficient.

\subsection{Stochastic Gradient Descent}
The stochastic gradient algorithm is often used in an LR model optimization problem owing to its fast convergence. Its learning rules are as follows:
\begin{equation}
    \begin{aligned}&\arg\min_{M,N}\varepsilon(M,N)\stackrel{SGD}{\Rightarrow}\\&\forall r_{uv}\in\Omega:\begin{cases}m_u^{(t)}\leftarrow m_u^{(t-1)}-\frac{\partial \varepsilon_{uv}\left ( M,N \right )  }{\partial m_u^{(t-1)}},\\
    n_v^{(t)}\leftarrow n_v^{(t-1)}-\frac{\partial \varepsilon_{uv}\left ( M,N \right )  }{\partial n_v^{(t-1)}}.\end{cases}\end{aligned}
\end{equation}

\begin{equation}
    \begin{aligned}
    =\begin{cases}m_u^{(t)}\leftarrow m_u^{(t-1)}+\eta\Big(e_{uv}n_v^{(t-1)}-\lambda m_u^{(t-1)}\Big),\\n_v^{(t)}\leftarrow n_v^{(t-1)}+\eta\Big(e_{uv}m_u^{(t-1)}-\lambda n_v^{(t-1)}\Big),\end{cases}\end{aligned}
\end{equation}
where $t$ denotes the $t$-th instance, $\eta$ is the learning rate and $e_{uv}=r_{ui} - \langle m_{u},n_{v} \rangle$.

\begin{figure}[t]
  \centering
  \includegraphics[width=0.5\linewidth]{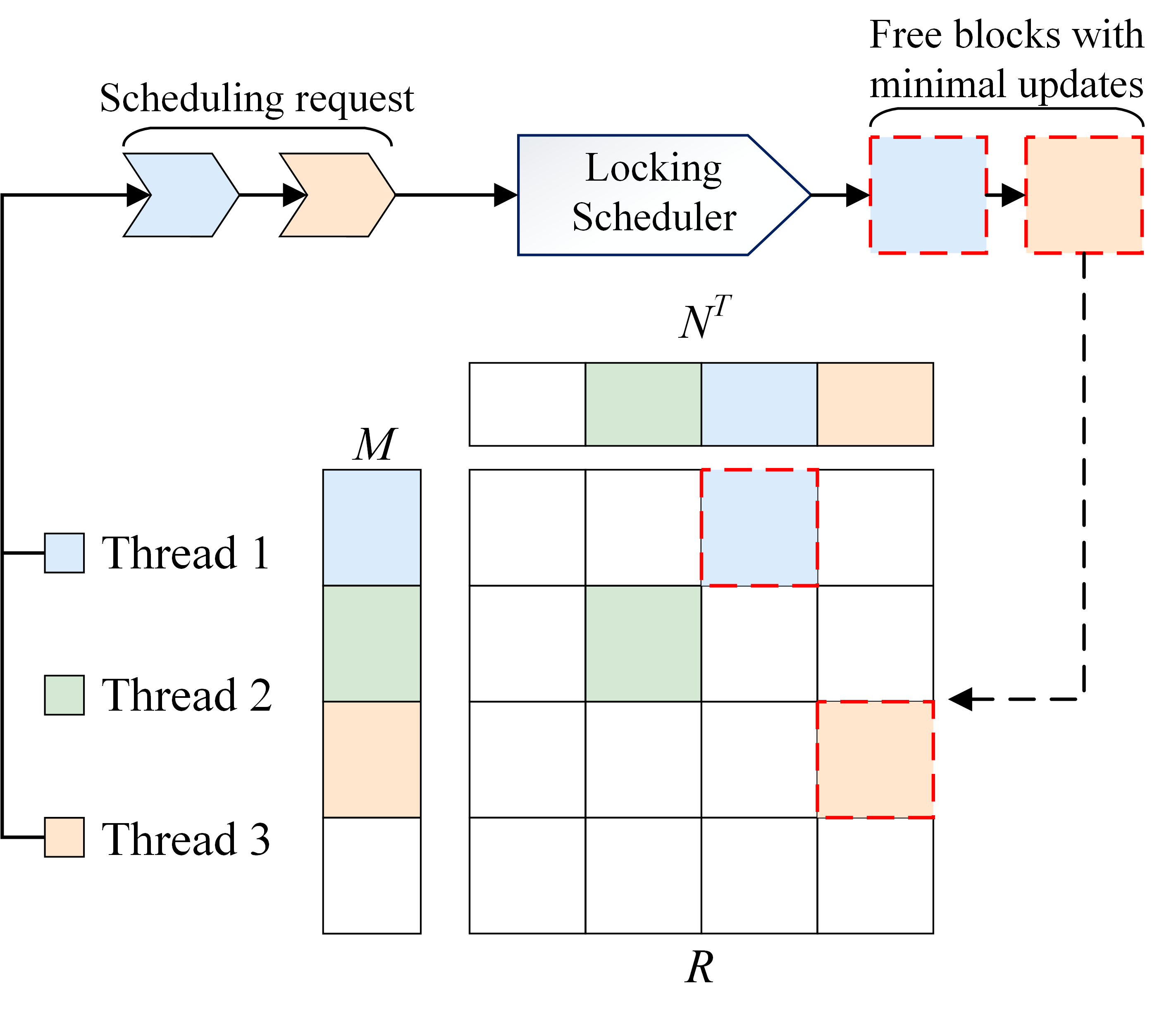}
  \caption{An illustration of the FPSGD scheduler with 3 threads. This matrix is divided into $4\times4$ sub-blocks. Threads 1 and 3 launch scheduling requests to this scheduler simultaneously, but the scheduler can only handle a request at a time due to the global lock of this scheduler.}
  \label{fig:1a}
\end{figure}

\section{The A$^{2}$PSGD-based LR Model}
\label{III}
\subsection{Asynchronous Dynamic Scheduling}
\label{III.A}
Following the principle of FPSGD~\cite{zhuang2013fast}, an asynchronous parallel SGD algorithm for HDS data LR, this paper blocks an HDS matrix into $(c+1)\times(c+1)$ matrix sub-blocks (where $c$ is the number of threads) and introduces the concept of a free block, representing a sub-block that shares neither the same rows nor the columns as the sub-blocks being concurrently processed. FPSGD implements asynchronous parallel optimization by scheduling free blocks with mininal updates to threads via a scheduler, where each thread updates only a currently scheduled sub-block. However, FPSGD's scalability is limited due to its scheduler relying on a global lock, causing queuing of multiple threads for lock acquisition.
Fig.~\ref{fig:1a} depicts an illustration of the FPSGD scheduler with 3 threads.

\begin{figure}[t]
  \centering
  \includegraphics[width=0.5\linewidth]{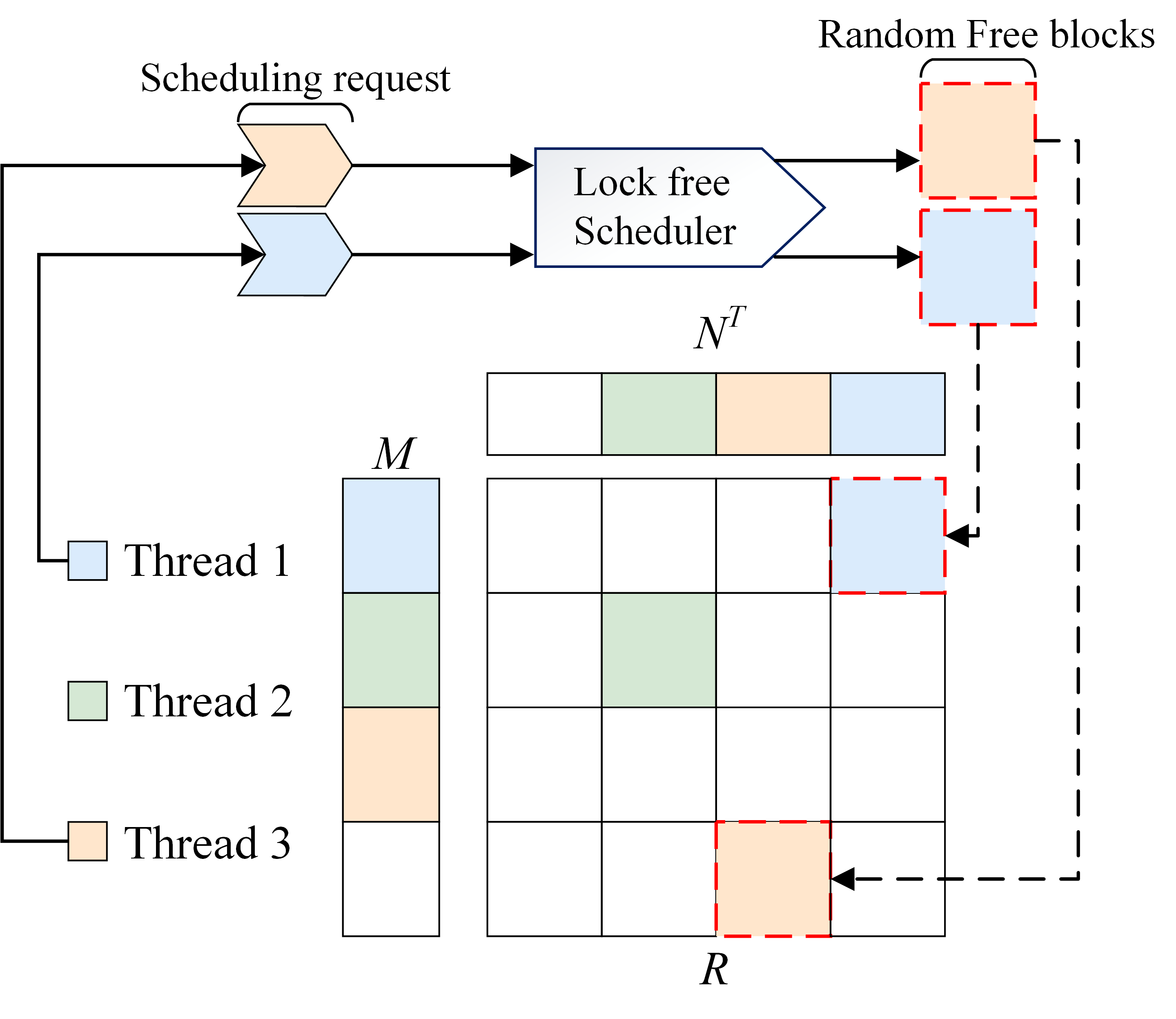}
  \caption{An illustration of the A$^{2}$PSGD scheduler with 3 threads. This matrix is divided into $4\times4$ sub-blocks. Threads 1 and 3 launch requests to the scheduler simultaneously, benefiting from A$^{2}$PSGD's lock-free scheduler, the scheduler can handle scheduling requests from multiple threads at a time.}
  \label{fig:1b}
\end{figure}

The A$^{2}$PSGD-based LR model adopts an improved scheduler~\cite{nishioka2015scalable}, which removes the global lock in the FPSGD's scheduler. The scheduler can respond to scheduling requests from multiple threads concurrently, addressing the poor scalability of FPSGD. Fig.~\ref{fig:1b} depicts an illustration of the A$^{2}$PSGD scheduler with 3 threads. It employs row and column locks to lock non-parallelizable rows and columns. A thread that launches a scheduling request to the scheduler randomly selects a row block index $rowBlockId$ and a column block index $colBlockId$ in the range $(1, c+1)$ and tries to acquire a lock for the $(rowBlockId, colBlockId)$ subblock. If the thread can acquire this lock, it treats the block as a free block and includes it in its update queue. Once the update is complete, the thread will perform an unlock operation on this block. However, the scheduler still suffers from the curse of the last reducer~\cite{oh2015fast}. This means that sub-blocks containing more instances are updated less frequently, thus affecting the convergence of the A$^{2}$PSGD-based LR model.

\begin{algorithm}[t]
\setcounter{algocf}{0}
 \caption{Load-balancing strategy-based blocking algorithm}
 \label{alg:2}
  \KwIn{An HDS matrix $R$, known instances $|\Omega|$, nodes $|U|$, nodes $|V|$ and threads $c$}
  \KwOut{sub-blocks $R_{ij}$ for $\forall i,j\in \left \{ 1,2,...,c+1 \right \}$  }
    Map$\left \langle int,int \right \rangle $ rowBlockMap\;
    Map$\left \langle int,int \right \rangle $ colBlockMap\;
  \tcp{\mycommentstyle{Row-wise blocking}}
  entriesPerRowBlock = $|\Omega| \setminus (c+1)$\;
  int rowBlockId = 0\;
  int rowEntriesCount = 0\;
    \For{$u = 1, \ldots, |U| $}
    { 
        numRowRatings = size of $r_{u,:}$\;
        rowEntriesCount += numRowRatings\;
        \If{rowEntriesCount $ \ge$ entriesPerRowBlock $\parallel u==|U|$} 
            { 
                Add $\left ( u+1, rowBlockId \right )$ to rowBlockMap\;
                rowEntriesCount = 0\;
                rowBlockId += 1\;
            }
    }
    \tcp{\mycommentstyle{Col-wise blocking}}
    entriesPerColBlock = $|\Omega| \setminus (c+1)$\;
    int colBlockId = 0\;
    int colEntriesCount = 0\;
    \For{$v = 1, \ldots, |V| $}
    { 
        numColRatings = size of $r_{:,v}$\;
        colEntriesCount += numcolRatings\;
        \If{colEntriesCount $ \ge$ entriesPercolBlock $\parallel v==|V|$} 
            { 
                Add $\left ( v+1, colBlockId \right )$ to colBlockMap\;
                colEntriesCount = 0\;
                colBlockId += 1\;
            }
    }
    \tcp{\mycommentstyle{Sub-block blocking}}
    Blocking $R$ into sub-blocks $R_{ij}$ for $\forall i,j\in \left \{ 1,2,...,c+1 \right \}$ accoding to rowBlockMap and colBlockMap\;
    return sub-blocks $R_{ij}$ for $\forall i,j\in \left \{ 1,2,...,c+1 \right \}$\; 
\end{algorithm}

\subsection{Load Balancing Strategy}
\label{III.B}
The FPSGD-based LR model blocks an HDS matrix $R$ into $(c+1)\times(c+1)$ sub-blocks, i.e., $c+1$ row blocks and $c+1$ col blocks. Accordingly, node set $U$ is divided into $(c+1)$ disjoint subsets $U^*= \left \{ U_1, U_2, ..., U_{c+1} \right \}$, node set $V$ is divided into $(c+1)$ disjoint subsets $ V^*=\left \{ V_1, V_2, ..., V_{c+1}  \right \}$.
Each sub-block satisfies $|U_{1}|=|U_{2}|=...=|U_{c+1}|=|U|/(c+1)$ and $|V_{1}|=|V_{2}|=...=|V_{c+1}|=|V|/(c+1)$. This method blocks an HDS matrix into equal-sized sub-blocks according to the number of nodes. However, it does not take into account the balanced distribution of instances across the sub-blocks. 
Compared with a sub-block containing fewer instances, the updated count of a sub-block containing more instances is fewer. This affects the convergence speed of the FPSGD-based LR model.

This section introduces a load balancing strategy for solving the problem. The strategy promotes the distribution of instances as evenly as possible across sub-blocks.

\subsubsection*{{\bf Definition 3} -- A Block}
A sub-block is denoted by $R_{ij}$ for $\forall i,j\in \left \{ 1,2,...,c+1 \right \}$, the $i$-th row block is denoted as $R_{i,:}$ and the $j$-th column block is denoted as $R_{:,j}$.

\subsubsection*{{\bf Definition 4} -- The number of instances contained in a block}
$\left \langle R_{ij} \right \rangle$ denotes the number of instances contained in a sub-block, $\left \langle R_{i,:} \right \rangle$ denotes the number of instances contained in the $i$-th row block and $\left \langle R_{:,j} \right \rangle$ denotes the number of instances contained in the $j$-th column block.

Unlike the equal-sized blocking method, the load-balancing strategy-based blocking method using a greedy algorithm blocks the node sets $U$ and $V$ into $U^\bot=\left \{ U_{1},U_{2},...,U_{c+1} \right \}$ and $V^\bot =\left \{ V_{1},V_{2},...,V_{c+1} \right \}$.
The corresponding each row block statisfy $\left \langle R_{1,:} \right \rangle \simeq\left \langle R_{2,:} \right \rangle \simeq...\simeq\left \langle R_{c+1,:} \right \rangle \simeq\Omega / (c+1)$ and $\left \langle R_{:,1} \right \rangle \simeq\left \langle R_{:,2} \right \rangle \simeq...\simeq\left \langle R_{:,c+1} \right \rangle \simeq\Omega / (c+1)$. With the guidance of $U^\bot$ and $V^\bot$, an HDS matrix is evenly blocked such that $R_{ij}$ for $\forall i,j\in \left \{ 1,2,...,c+1 \right \}$ is as close to $\Omega / (c+1)^2$ as possible. Algorithm \textcolor{red}{\ref{alg:2}} details the load-balancing strategy-based blocking method.

\subsection{Accelerated Optimization Scheme in A$^{2}$PSGD}
\label{III.C}
To further accelerate the convergence of A$^{2}$PSGD, NAG~\cite{qu2019accelerated} is incorporated into the learning scheme of A$^{2}$PSGD. NAG, a variant of the momentum method, is designed to enhance SGD convergence stability and accelerate convergence. The momentum method introduces momentum terms, representing the direction and speed of the previous move, to mitigate oscillations and wobbles in gradient descent. However, it may cause excessive movement when approaching the minimum value. The NAG-based accelerated optimization scheme for A$^{2}$PSGD addresses this issue by performing an additional update to the previous position before computing the gradient. Its learning scheme is as follows:
\begin{equation}
    \begin{cases}
    \phi_{u}^{(t)}=\gamma \phi_{u}^{(t-1)}-\eta \frac{\partial \varepsilon_{u i}\left(m_{u}^{(t-1)}+\gamma \phi_{u}^{(t-1)}, N\right)}{\partial m_{u}^{(t-1)}}, \\
    m_{u}^{(t)}=m_{u}^{(t-1)}+\phi_{u}^{(t)},
    \end{cases}
\end{equation}

\begin{equation}
    \begin{cases}
    \varphi_v^{(t)}=\gamma\varphi_v^{(t-1)}-\eta\frac{\partial\varepsilon_{uv}\left(M,n_v^{(t-1)}+\gamma\varphi_v^{(t-1)}\right)}{\partial n_v^{(t-1)}},\\
    n_v^{(m)}=n_v^{(t-1)}+\varphi_v^{(t)},\end{cases}
\end{equation}
where $\phi_{u}$ is the momentum vector of $m_u$, $\varphi_v$ is the momentum vector of $n_v$, $\phi_u^{(0)}$ is the initial state of $\phi_{u}$, $\varphi_v^{(0)}$ is the initial state of $\phi_{u}$ and $\gamma$ is the momentum coefficient, respectively.

Before training, A$^{2}$PSGD’s accelerated optimization scheme initializes momentum matrices $\phi^{|U|\times{|D|}}=0$ and $\varphi^{|V|\times{|D|}}=0$. At the $t$-th instance, it updates the momentum vector $\phi_{u}^{(t)}/\varphi_v^{(t)}$ via the momentum vector $\phi_{u}^{(t-1)}/\varphi_v^{(t-1)}$ on the $(t-1)$-th instance and computes the gradient at this predicted position to update feature parameters $m_u^{(t)}/n_v^{(t)}$.

An HDS matrix is balancedly divided into $(c+1)\times(c+1)$ sub-blocks by using the load balancing strategy-based blocking algorithm in Section~\ref{III.A}. A free block is scheduled to all threads by the asynchronous dynamic scheduling method in Section~\ref{III.B}. Finally, the LR model parameters are optimized by using the accelerated optimization scheme in Section~\ref{III.C}. By doing so, the A$^{2}$PSGD-based LR model is proposed.

\vspace{-15pt}
\begin{table}[bp]
\centering
\caption{The hypermeters on Movielens 1M dataset}
\label{table2}
\begin{tabular}{c|ccccc}
\hline
Hyperparameter & Hogwild! & DSGD & ASGD & FPSGD & A$^{2}$PSGD \\ \hline
$\lambda$              & 3e-2        & 3e-2    & 3e-2    & 3e-2     & 5e-2      \\
$\eta$              & 6e-4        & 6e-4    & 6e-4    & 6e-4     & 1e-4      \\
$\gamma$              & -        & -    & -    & -     & 9e-1      \\ \hline
\end{tabular}
\end{table}

\begin{table}[bp]
\centering
\caption{The hypermeters on Epinion 665K dataset}
\label{table3}
\begin{tabular}{c|ccccc}
\hline
Hyperparameter & Hogwild! & DSGD & ASGD & FPSGD & A$^{2}$PSGD \\ \hline
$\lambda$              & 5e-1        & 5e-1    & 5e-1    & 5e-1     & 4e-1      \\
$\eta$              & 2e-3        & 2e-3    & 2e-3    & 2e-3     & 2e-4      \\
$\gamma$              & -        & -    & -    & -     & 9e-1      \\ \hline
\end{tabular}
\end{table}

\vspace{10pt}
\section{Experiments}
\label{IV}
\subsection{Settings}
\subsubsection{Datasets}
The datasets used in the empirical studies are derived from two different industrial application scenarios:
\begin{itemize}
    \item Movielens 1M~\cite{li2019mixture}. The dataset is provided by the GroupLens Research Project at the University of Minnesota. It contains 1,000,209 anonymous ratings of approximately 3,706 movies made by 6,040 users.
    \item Epionion 665K~\cite{ijcai2019-191}. The dataset comes from the Epinions.com website, a consumer review platform where users can review products and rate the reviews of others. It contains 664,824 ratings given by 40,163 users to 139,738 to items.
\end{itemize}

We randomly divided them into training and test sets with 70\% and 30\% respectively.

\subsubsection{Baseline models}
\begin{itemize}
    \item Hogwild!~\cite{recht2011hogwild}: Asynchronous SGD without considering overwriting problems.
    \item DSGD~\cite{gemulla2011large}: Distributed SGD with bulk synchronizations.
    \item ASGD~\cite{luo2012parallel}: Alternating SGD with decoupled parameter parallelism.
    \item FPSGD~\cite{zhuang2013fast}: Asynchronous Parallel SGD with a scheduler having a shared lock.
\end{itemize}

The above optimization algorithms are dedicated to HDS data and based on them the baseline parallel LR models are constructed.

\subsubsection{Experimental environment}
All experiments are conducted on a server equipped with two Intel Xeon Gold 6230R CPUs (26 cores, 2.1 GHz) and 1 TB RAM. All models are implemented in C++.

\subsubsection{Evaluation metrics}
HDS data LR is essentially a missing data prediction problem~\cite{9238448, 9152087, 9072622, 9462337, 9785520, 9357412}. The commonly-used root mean squared error (RMSE) and mean absolute error (MAE) are adopted to evaluate the prediction accuracy on the test set, defined as follows:

\begin{equation*}
    RMSE=\sqrt{\left(\sum_{r_{uv}\in\Psi}\left(r_{u,i}-\hat{r}_{uv}\right)^2\right)\Big/|\Psi|},     
    MAE=\Bigg(\sum_{r_{uv}\in\Psi}\Big|r_{uv}-\hat{r}_{uv}\Big|_{abs}\Bigg)\Big/|\Psi|,
\end{equation*}
where $\Psi$ denotes the test set.

\subsubsection{Hyperparameters}
The hyperparameter settings for the A$^{2}$PSGD-based LR model and the baseline LR models on the two datasets are shown in Tables~\ref{table2} and~\ref{table3}. Note that the settings of hyperparameters $\lambda$ and $\eta$ are obtained by performing grid search and ten-fold cross-validation on the validation set additionally divided on the test set $\Psi$.

\subsection{Results}
In this section, we compare the proposed A$^{2}$PSGD-based LR model with the baseline LR models in terms of prediction accuracy and training time. Figs.~\ref{Fig2} and \ref{Fig3} depict the convergence curves for all models at 32 threads, and Tables \textcolor{red}{\ref{table4}} and \textcolor{red}{\ref{table5}} list the prediction error and training time of all models at 32 threads, respectively. Based on the experimental results, the following is discussed.

\begin{table*}[tp]
\centering
\caption{Prediction accuracy for all models at 32 threads on Movielens 1M and Epinion 665K datasets}
\label{table4}
\begin{tabular}{lllllll}
\hline
Dataset                                            & Case & Hogwild! & DSGD & ASGD & FPSGD & A$^{2}$PSGD \\ \hline
\multicolumn{1}{l|}{\multirow{2}{*}{Movielens 1M}} & RMSE & 0.8602±1.43e-4 & 0.8583±0 & 0.8584±0 & 0.8585±2.04e-05 & \textbf{0.8552±6.78e-05} \\
\multicolumn{1}{l|}{}                              & MAE  & 0.6761±1.23e-4 & 0.6746±0 & 0.6747±1.11e-16 & 0.6747±1.58e-05 & \textbf{0.6741±1.63e-04} \\ \hline
\multicolumn{1}{l|}{\multirow{2}{*}{Epinion 665K}} & RMSE & 2.0239±3.19e-4 & 2.0266±0 & 2.0264±0 & 2.0316±7.13e-04 & \textbf{2.0165±1.43e-04} \\
\multicolumn{1}{l|}{}                              & MAE  & 1.4910±3.67e-4 & 1.4956±0 & 1.4953±0 & 1.5037±9.92e-04 & \textbf{1.4705±2.35e-04} \\ \hline
\end{tabular}
\end{table*}

\begin{table*}[tp]
\centering
\caption{Training time for all models at 32 threads on Movielens 1M and Epinion 665K datasets}
\label{table5}
\begin{tabular}{lllllll}
\hline
Dataset                                            & Case & Hogwild! & DSGD & ASGD & FPSGD & A$^{2}$PSGD \\ \hline
\multicolumn{1}{l|}{\multirow{2}{*}{Movielens 1M}} & RMSE-time & 15.18±0.02 & 23.11±3.21 & 18.00±0.92 & 322.16±9.86 & \textbf{14.98±0.12} \\
\multicolumn{1}{l|}{}                              & MAE-time  & 17.34±1.01 &    25.28±3.59 & 19.67±0.92 & 336.26±14.71 & \textbf{17.13±1.43} \\ \hline
\multicolumn{1}{l|}{\multirow{2}{*}{Epinion 665K}} & RMSE-time & 38.45±7.02 & 74.10±1.50 & 78.70±5.36 & 519.06±26.79 & \textbf{38.34±4.85} \\
\multicolumn{1}{l|}{}                              & MAE-time  & 46.58±1.69 & 91.15±2.48 & 96.72±6.47 & 519.06±26.79 & \textbf{36.64±4.75} \\ \hline
\end{tabular}
\end{table*}

\subsubsection{Prediction Accuracy}
\begin{figure}[htbp]
    \centering
    \subfigure[RMSE convergence curve on Movielens 1M dataset]{
        \centering
        \includegraphics[width=6cm]{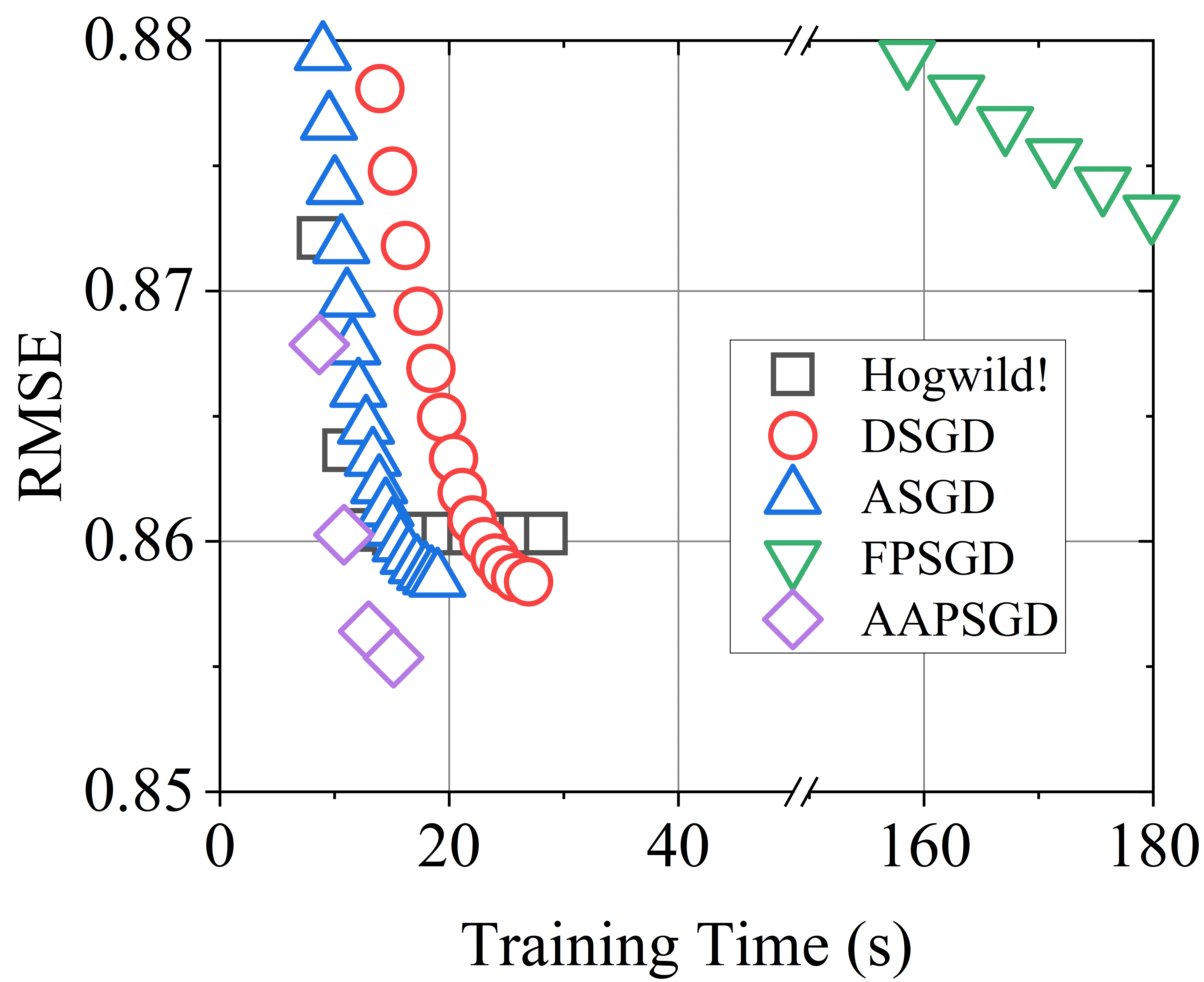}
        \label{Fig2a}
    }
    \subfigure[RMSE convergence curve on Epinion 665K dataset]{
        \centering
        \includegraphics[width=6cm]{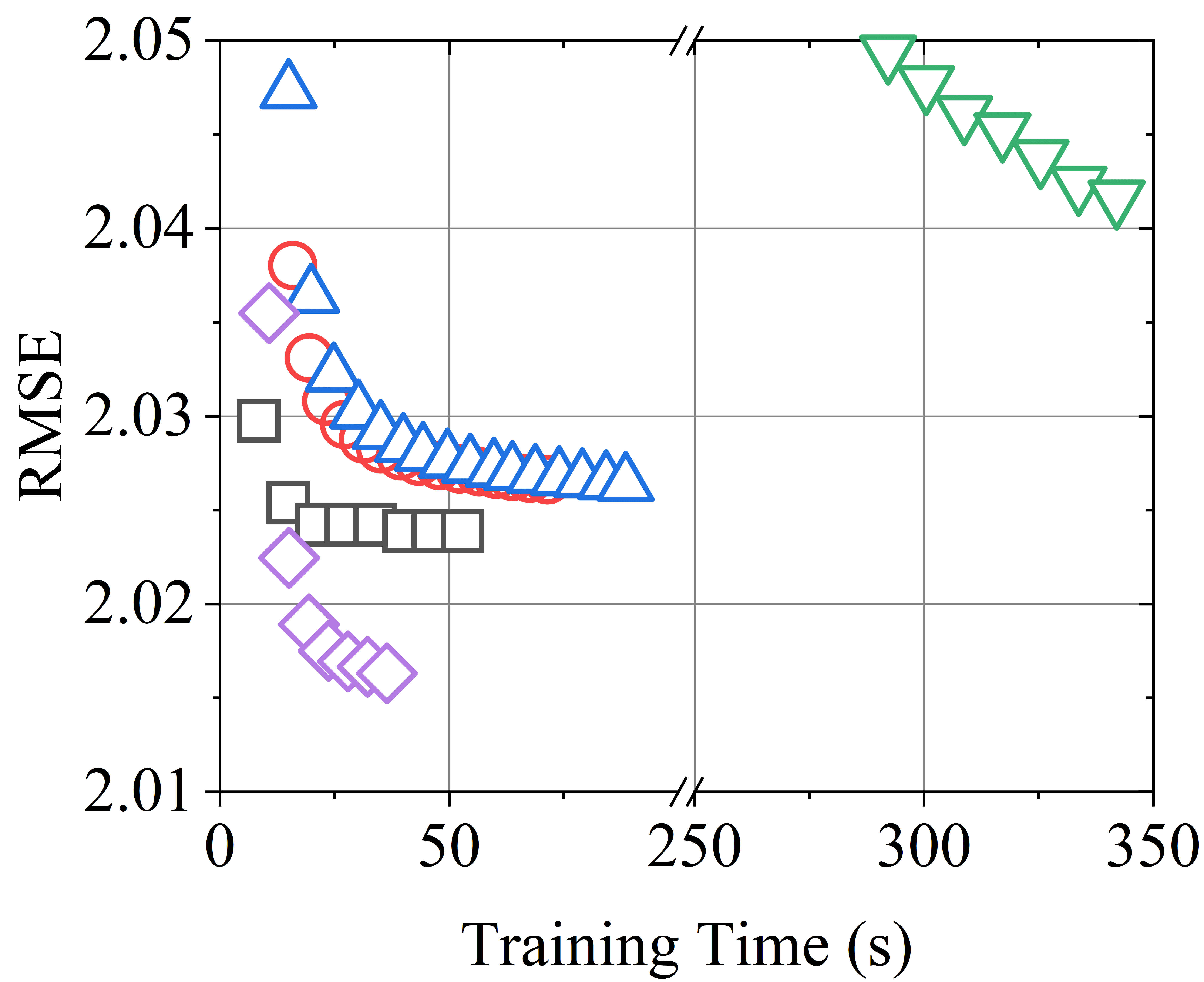}
        \label{Fig2b}
    }
    \caption{RMSE convergence curves for all models at 32 threads. All subfigures share the same legend.}
    \label{Fig2}
\end{figure}

\textbf{The prediction accuracy of the proposed A$^{2}$PSGD-based LR model is the highest.} As shown in Fig.~\ref{Fig2} and Table~\ref{table4}, A$^{2}$PSGD-based LR model has the lowest RMSE and MAE compared to the benchmark models on both datasets. For instance, on the Movielens 1M dataset, the RMSE of the A$^{2}$PSGD-based LR model is 0.8522, which is lower than the second best DSGD-based LR model by 0.7\%. It is also slightly lower than 0.8602, 0.8584 and 0.8585 of the Hogwild!, ASGD and FPSGD-based LR models, respectively. The improvement in the prediction accuracy of the A$^{2}$PSGD-based LR model can be attributed to the fact that the learning scheme incorporating NAD suppresses oscillations generated near saddle points or local minima.

\subsubsection{Training time}
\begin{figure}[htbp]
        \centering
	\subfigure[MAE convergence curve on Movielens 1M dataset]{
		\includegraphics[width=6cm]{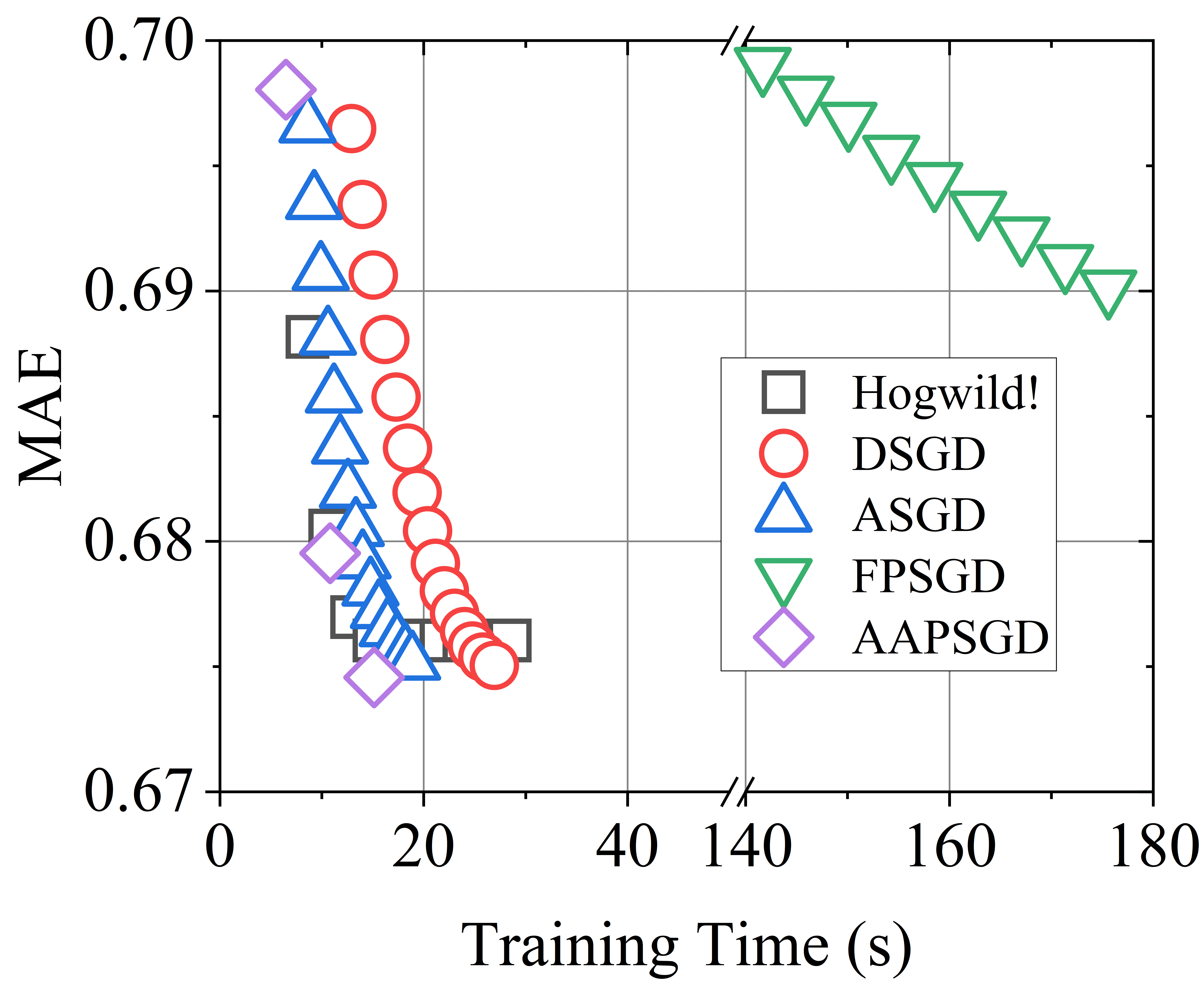}
		\label{Fig3a}
	}
	\subfigure[MAE convergence curve on Epinion 665K dataset]{
        \centering
		\includegraphics[width=6cm]{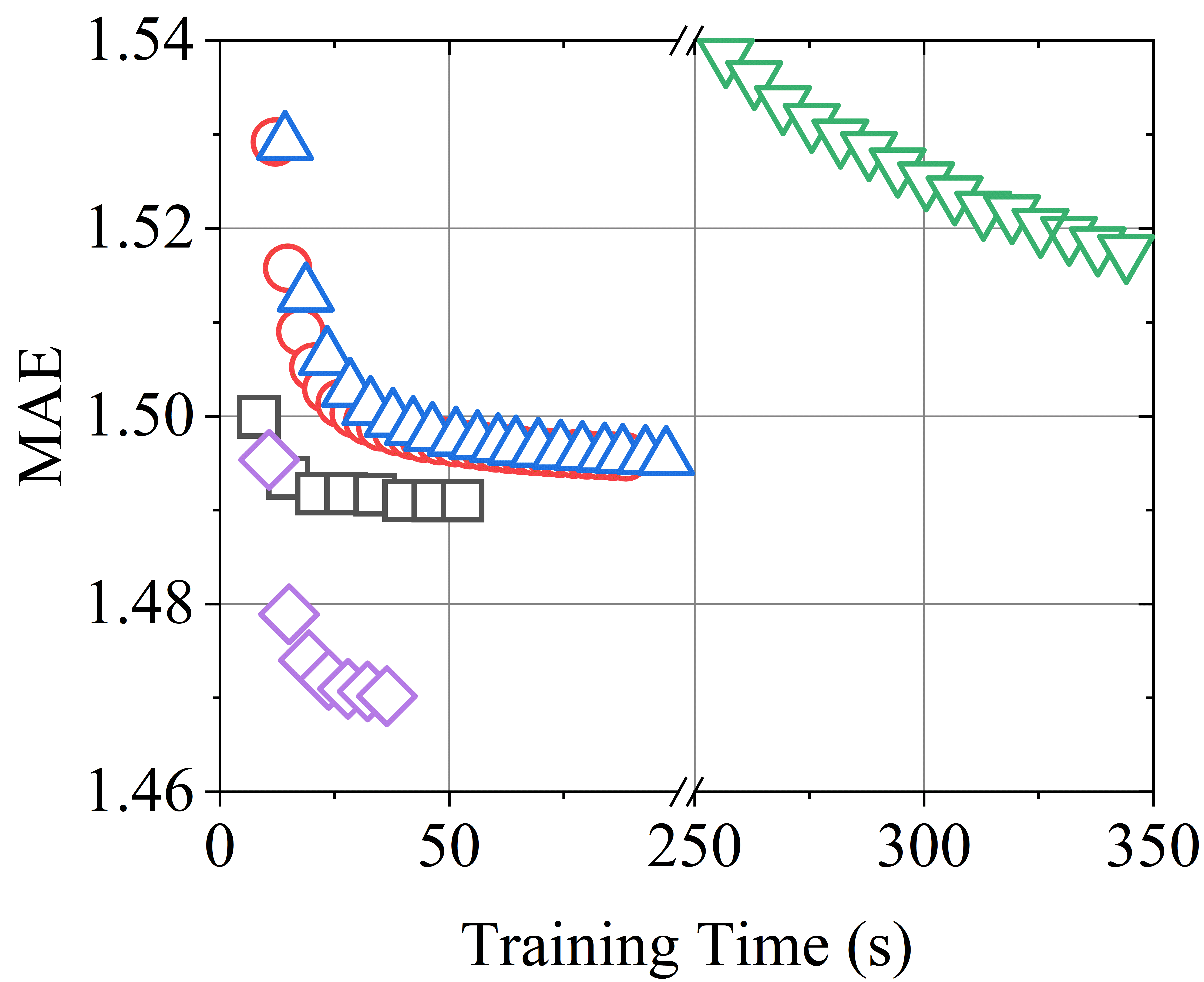}
		\label{Fig3b}
	}
	\caption{MAE convergence curves for all models at 32 threads. All subfigures share the same legend.}\label{Fig3}
\end{figure}

\textbf{The computational efficiency of the A$^{2}$PSGD-based LR model is the highest}. As shown in Fig.~\ref{Fig3} and Table~\ref{table5}, the A$^{2}$PSGD-based LR model has a lower training time than the baseline parallel LR models on both datasets. For instance, on the Epinion 665K dataset, the RMSE training time of the A$^{2}$PSGD-based LR model is slightly lower than that of the Hogwild!-based LR model, at 38.34 and 38.45, respectively. However, Hogwild! suffers from the update overwriting problems, which reaches a worse local minimum at the termination iteration.

In summary,  benefiting from the asynchronous dynamic scheduling, load balancing strategy, and accelerated optimization scheme, the A$^{2}$PSGD-based LR model proposed in this paper can achieve better gains than the existing state-of-the-art parallel LR models both in terms of prediction accuracy and computational efficiency.

\vspace{-10pt}

\section{Conclusions}
\label{V}
In this paper, we propose the A$^{2}$PSGD-based LR model, removing the global lock in the scheduler of an existing parallel LR model via asynchronous dynamic scheduling. This allows the scheduler to handle requests from multiple threads and balancedly blocks an HDS matrix via a load-balancing strategy, which equalizes the number of updates to each sub-block and promotes the convergence speed of the LR model. Additionally, Nesterov accelerated gradient is incorporated into the learning scheme of A$^{2}$PSGD to suppress optimizer oscillations around local minima and accelerate convergence. Experimental results demonstrate the model's efficiency in representing large-scale HDS matrices with higher prediction accuracy and computational efficiency. Future work will extend this model to address the time-varying properties of HDS data by considering higher-order tensors.

\bibliographystyle{IEEEtran}
\bibliography{ref}

\begin{thebibliography}{10}
\providecommand{\url}[1]{#1}
\csname url@samestyle\endcsname
\providecommand{\newblock}{\relax}
\providecommand{\bibinfo}[2]{#2}
\providecommand{\BIBentrySTDinterwordspacing}{\spaceskip=0pt\relax}
\providecommand{\BIBentryALTinterwordstretchfactor}{4}
\providecommand{\BIBentryALTinterwordspacing}{\spaceskip=\fontdimen2\font plus
\BIBentryALTinterwordstretchfactor\fontdimen3\font minus \fontdimen4\font\relax}
\providecommand{\BIBforeignlanguage}[2]{{%
\expandafter\ifx\csname l@#1\endcsname\relax
\typeout{** WARNING: IEEEtran.bst: No hyphenation pattern has been}%
\typeout{** loaded for the language `#1'. Using the pattern for}%
\typeout{** the default language instead.}%
\else
\language=\csname l@#1\endcsname
\fi
#2}}
\providecommand{\BIBdecl}{\relax}
\BIBdecl

\bibitem{Luo15}
X.~Luo, Y.~Zhou, Z.~Liu, and M.~Zhou, ``Fast and accurate non-negative latent factor analysis of high-dimensional and sparse matrices in recommender systems,'' \emph{IEEE Transactions on Knowledge and Data Engineering}, vol.~35, no.~4, pp. 3897--3911, 2023.

\bibitem{Zhong5}
Y.~Zhong, K.~Liu, S.~Gao, and X.~Luo, ``Alternating-direction-method of multipliers-based adaptive nonnegative latent factor analysis,'' \emph{IEEE Transactions on Emerging Topics in Computational Intelligence}, pp. 1--15, 2024.

\bibitem{Wang7}
J.~Wang, W.~Li, and X.~Luo, ``A distributed adaptive second-order latent factor analysis model,'' \emph{IEEE/CAA Journal of Automatica Sinica}, pp. 1--3, 2024.

\bibitem{Luo16}
X.~Luo, Y.~Zhong, Z.~Wang, and M.~Li, ``An alternating-direction-method of multipliers-incorporated approach to symmetric non-negative latent factor analysis,'' \emph{IEEE Transactions on Neural Networks and Learning Systems}, vol.~34, no.~8, pp. 4826--4840, 2023.

\bibitem{Wu13}
D.~Wu, X.~Luo, Y.~He, and M.~Zhou, ``A prediction-sampling-based multilayer-structured latent factor model for accurate representation to high-dimensional and sparse data,'' \emph{IEEE Transactions on Neural Networks and Learning Systems}, vol.~35, no.~3, pp. 3845--3858, 2024.

\bibitem{ray2021various}
P.~Ray, S.~S. Reddy, and T.~Banerjee, ``Various dimension reduction techniques for high dimensional data analysis: a review,'' \emph{Artificial Intelligence Review}, vol.~54, no.~5, pp. 3473--3515, 2021.

\bibitem{Tugnait}
J.~K. Tugnait, ``Learning sparse high-dimensional matrix-valued graphical models from dependent data,'' \emph{IEEE Transactions on Signal Processing}, vol.~72, pp. 3363--3379, 2024.

\bibitem{Luo17}
X.~Luo, L.~Wang, P.~Hu, and L.~Hu, ``Predicting protein-protein interactions using sequence and network information via variational graph autoencoder,'' \emph{IEEE/ACM Transactions on Computational Biology and Bioinformatics}, vol.~20, no.~5, pp. 3182--3194, 2023.

\bibitem{Bi18}
F.~Bi, X.~Luo, B.~Shen, H.~Dong, and Z.~Wang, ``Proximal alternating-direction-method-of-multipliers-incorporated nonnegative latent factor analysis,'' \emph{IEEE/CAA Journal of Automatica Sinica}, vol.~10, no.~6, pp. 1388--1406, 2023.

\bibitem{liu2023constraint}
Z.~Liu, X.~Luo, Z.~Wang, and X.~Liu, ``Constraint-induced symmetric nonnegative matrix factorization for accurate community detection,'' \emph{Information Fusion}, vol.~89, pp. 588--602, 2023.

\bibitem{9779537}
Q.~Wang, X.~Liu, T.~Shang, Z.~Liu, H.~Yang, and X.~Luo, ``Multi-constrained embedding for accurate community detection on undirected networks,'' \emph{IEEE Transactions on Network Science and Engineering}, vol.~9, no.~5, pp. 3675--3690, 2022.

\bibitem{10012357}
Z.~Liu, Y.~Yi, and X.~Luo, ``A high-order proximity-incorporated nonnegative matrix factorization-based community detector,'' \emph{IEEE Transactions on Emerging Topics in Computational Intelligence}, vol.~7, no.~3, pp. 700--714, 2023.

\bibitem{9340571}
X.~Luo, Z.~Liu, L.~Jin, Y.~Zhou, and M.~Zhou, ``Symmetric nonnegative matrix factorization-based community detection models and their convergence analysis,'' \emph{IEEE Transactions on Neural Networks and Learning Systems}, vol.~33, no.~3, pp. 1203--1215, 2022.

\bibitem{9865020}
Z.~Liu, G.~Yuan, and X.~Luo, ``Symmetry and nonnegativity-constrained matrix factorization for community detection,'' \emph{IEEE/CAA Journal of Automatica Sinica}, vol.~9, no.~9, pp. 1691--1693, 2022.

\bibitem{wu2022graph}
S.~Wu, F.~Sun, W.~Zhang, X.~Xie, and B.~Cui, ``Graph neural networks in recommender systems: a survey,'' \emph{ACM Computing Surveys}, vol.~55, no.~5, pp. 1--37, 2022.

\bibitem{chen2023bias}
J.~Chen, H.~Dong, X.~Wang, F.~Feng, M.~Wang, and X.~He, ``Bias and debias in recommender system: A survey and future directions,'' \emph{ACM Transactions on Information Systems}, vol.~41, no.~3, pp. 1--39, 2023.

\bibitem{xin2019non}
L.~Xin, Y.~Yuan, M.~Zhou, Z.~Liu, and M.~Shang, ``Non-negative latent factor model based on $\beta$-divergence for recommender systems,'' \emph{IEEE Transactions on Systems, Man, and Cybernetics: Systems}, vol.~51, no.~8, pp. 4612--4623, 2019.

\bibitem{cui2020personalized}
Z.~Cui, X.~Xu, X.~Fei, X.~Cai, Y.~Cao, W.~Zhang, and J.~Chen, ``Personalized recommendation system based on collaborative filtering for iot scenarios,'' \emph{IEEE Transactions on Services Computing}, vol.~13, no.~4, pp. 685--695, 2020.

\bibitem{9411671}
D.~Wu, M.~Shang, X.~Luo, and Z.~Wang, ``An l1-and-l2-norm-oriented latent factor model for recommender systems,'' \emph{IEEE Transactions on Neural Networks and Learning Systems}, vol.~33, no.~10, pp. 5775--5788, 2022.

\bibitem{gao2024causal}
C.~Gao, Y.~Zheng, W.~Wang, F.~Feng, X.~He, and Y.~Li, ``Causal inference in recommender systems: A survey and future directions,'' \emph{ACM Transactions on Information Systems}, vol.~42, no.~4, pp. 1--32, 2024.

\bibitem{hu2021distributed}
L.~Hu, S.~Yang, X.~Luo, H.~Yuan, K.~Sedraoui, and M.~Zhou, ``A distributed framework for large-scale protein-protein interaction data analysis and prediction using mapreduce,'' \emph{IEEE/CAA Journal of Automatica Sinica}, vol.~9, no.~1, pp. 160--172, 2021.

\bibitem{poleksic2023hyperbolic}
A.~Poleksic, ``Hyperbolic matrix factorization improves prediction of drug-target associations,'' \emph{Scientific Reports}, vol.~13, no.~1, p. 959, 2023.

\bibitem{zhang2023graph}
J.~Zhang and M.~Xie, ``Graph regularized non-negative matrix factorization with l 2, 1 norm regularization terms for drug--target interactions prediction,'' \emph{BMC bioinformatics}, vol.~24, no.~1, p. 375, 2023.

\bibitem{10164211}
Z.~Cui, S.-G. Wang, Y.~He, Z.-H. Chen, and Q.-H. Zhang, ``Deeptppred: A deep learning approach with matrix factorization for predicting therapeutic peptides by integrating length information,'' \emph{IEEE Journal of Biomedical and Health Informatics}, vol.~27, no.~9, pp. 4611--4622, 2023.

\bibitem{ai2023low}
C.~Ai, H.~Yang, Y.~Ding, J.~Tang, and F.~Guo, ``Low rank matrix factorization algorithm based on multi-graph regularization for detecting drug-disease association,'' \emph{IEEE/ACM Transactions on Computational Biology and Bioinformatics}, vol.~20, no.~5, pp. 3033--3043, 2023.

\bibitem{gao2023predicting}
H.~Gao, J.~Sun, Y.~Wang, Y.~Lu, L.~Liu, Q.~Zhao, and J.~Shuai, ``Predicting metabolite--disease associations based on auto-encoder and non-negative matrix factorization,'' \emph{Briefings in bioinformatics}, vol.~24, no.~5, p. bbad259, 2023.

\bibitem{shen2023matrix}
X.~Shen, H.~Liu, J.~Nie, and X.~Zhou, ``Matrix factorization with framelet and saliency priors for hyperspectral anomaly detection,'' \emph{IEEE Transactions on Geoscience and Remote Sensing}, vol.~61, pp. 1--13, 2023.

\bibitem{10173648}
C.~Cui, X.~Wang, S.~Wang, L.~Zhang, and Y.~Zhong, ``Unrolling nonnegative matrix factorization with group sparsity for blind hyperspectral unmixing,'' \emph{IEEE Transactions on Geoscience and Remote Sensing}, vol.~61, pp. 1--12, 2023.

\bibitem{10271326}
X.~Chen, W.~Xia, Z.~Yang, H.~Chen, Y.~Liu, J.~Zhou, Z.~Wang, Y.~Chen, B.~Wen, and Y.~Zhang, ``Soul-net: A sparse and low-rank unrolling network for spectral ct image reconstruction,'' \emph{IEEE Transactions on Neural Networks and Learning Systems}, pp. 1--15, 2023.

\bibitem{10004837}
Z.~Zha, B.~Wen, X.~Yuan, S.~Ravishankar, J.~Zhou, and C.~Zhu, ``Learning nonlocal sparse and low-rank models for image compressive sensing: Nonlocal sparse and low-rank modeling,'' \emph{IEEE Signal Processing Magazine}, vol.~40, no.~1, pp. 32--44, 2023.

\bibitem{Xie_2023_ICCV}
Y.~Xie, C.~Xu, M.-J. Rakotosaona, P.~Rim, F.~Tombari, K.~Keutzer, M.~Tomizuka, and W.~Zhan, ``Sparsefusion: Fusing multi-modal sparse representations for multi-sensor 3d object detection,'' in \emph{Proceedings of the IEEE/CVF International Conference on Computer Vision (ICCV)}, October 2023, pp. 17\,591--17\,602.

\bibitem{he2023improved}
Y.~He, Y.~Su, X.~Wang, J.~Yu, and Y.~Luo, ``An improved method mss-yolov5 for object detection with balancing speed-accuracy,'' \emph{Frontiers in Physics}, vol.~10, p. 1101923, 2023.

\bibitem{luo2019temporal}
X.~Luo, H.~Wu, H.~Yuan, and M.~Zhou, ``Temporal pattern-aware qos prediction via biased non-negative latent factorization of tensors,'' \emph{IEEE transactions on cybernetics}, vol.~50, no.~5, pp. 1798--1809, 2019.

\bibitem{qin2023adaptively}
W.~Qin, X.~Luo, and M.~Zhou, ``Adaptively-accelerated parallel stochastic gradient descent for high-dimensional and incomplete data representation learning,'' \emph{IEEE Transactions on Big Data}, vol.~10, no.~1, pp. 92--107, 2023.

\bibitem{luo2021novel}
X.~Luo, H.~Wu, Z.~Wang, J.~Wang, and D.~Meng, ``A novel approach to large-scale dynamically weighted directed network representation,'' \emph{IEEE Transactions on Pattern Analysis and Machine Intelligence}, vol.~44, no.~12, pp. 9756--9773, 2021.

\bibitem{qin2023parallel}
W.~Qin, X.~Luo, S.~Li, and M.~Zhou, ``Parallel adaptive stochastic gradient descent algorithms for latent factor analysis of high-dimensional and incomplete industrial data,'' \emph{IEEE Transactions on Automation Science and Engineering}, pp. 1--14, 2023.

\bibitem{9659145}
H.~Wu, X.~Luo, and M.~Zhou, ``Neural latent factorization of tensors for dynamically weighted directed networks analysis,'' in \emph{2021 IEEE International Conference on Systems, Man, and Cybernetics (SMC)}, Melbourne, Australia, 2021, pp. 3061--3066.

\bibitem{10555245}
J.~Chen, Y.~Yuan, and X.~Luo, ``Sdgnn: Symmetry-preserving dual-stream graph neural networks,'' \emph{IEEE/CAA Journal of Automatica Sinica}, vol.~11, no.~7, pp. 1717--1719, 2024.

\bibitem{10035508}
F.~Bi, T.~He, Y.~Xie, and X.~Luo, ``Two-stream graph convolutional network-incorporated latent feature analysis,'' \emph{IEEE Transactions on Services Computing}, vol.~16, no.~4, pp. 3027--3042, 2023.

\bibitem{10179251}
D.~Wu, Y.~He, and X.~Luo, ``A graph-incorporated latent factor analysis model for high-dimensional and sparse data,'' \emph{IEEE Transactions on Emerging Topics in Computing}, vol.~11, no.~4, pp. 907--917, 2023.

\bibitem{9932678}
M.~Chen, C.~He, and X.~Luo, ``Mnl: A highly-efficient model for large-scale dynamic weighted directed network representation,'' \emph{IEEE Transactions on Big Data}, vol.~9, no.~3, pp. 889--903, 2023.

\bibitem{9551506}
H.~Wu, X.~Luo, and M.~Zhou, ``Discovering hidden pattern in large-scale dynamically weighted directed network via latent factorization of tensors,'' in \emph{2021 IEEE 17th International Conference on Automation Science and Engineering (CASE)}, Lyon, France, 2021, pp. 1533--1538.

\bibitem{9783168}
D.~Wu, P.~Zhang, Y.~He, and X.~Luo, ``A double-space and double-norm ensembled latent factor model for highly accurate web service qos prediction,'' \emph{IEEE Transactions on Services Computing}, vol.~16, no.~2, pp. 802--814, 2023.

\bibitem{9839318}
Y.~Yuan, X.~Luo, M.~Shang, and Z.~Wang, ``A kalman-filter-incorporated latent factor analysis model for temporally dynamic sparse data,'' \emph{IEEE Transactions on Cybernetics}, vol.~53, no.~9, pp. 5788--5801, 2023.

\bibitem{9894115}
W.~Li, R.~Wang, X.~Luo, and M.~Zhou, ``A second-order symmetric non-negative latent factor model for undirected weighted network representation,'' \emph{IEEE Transactions on Network Science and Engineering}, vol.~10, no.~2, pp. 606--618, 2023.

\bibitem{10113599}
Y.~Zhou, X.~Luo, and M.~Zhou, ``Cryptocurrency transaction network embedding from static and dynamic perspectives: An overview,'' \emph{IEEE/CAA Journal of Automatica Sinica}, vol.~10, no.~5, pp. 1105--1121, 2023.

\bibitem{yuan2024fuzzy}
Y.~Yuan, J.~Li, and X.~Luo, ``A fuzzy pid-incorporated stochastic gradient descent algorithm for fast and accurate latent factor analysis,'' \emph{IEEE Transactions on Fuzzy Systems}, pp. 1--12, 2024.

\bibitem{chen2024generalized}
M.~Chen, R.~Wang, Y.~Qiao, and X.~Luo, ``A generalized nesterov's accelerated gradient-incorporated non-negative latent-factorization-of-tensors model for efficient representation to dynamic qos data,'' \emph{IEEE Transactions on Emerging Topics in Computational Intelligence}, pp. 1--15, 2024.

\bibitem{xue2021spatial}
J.~Xue, Y.-Q. Zhao, Y.~Bu, W.~Liao, J.~C.-W. Chan, and W.~Philips, ``Spatial-spectral structured sparse low-rank representation for hyperspectral image super-resolution,'' \emph{IEEE Transactions on Image Processing}, vol.~30, pp. 3084--3097, 2021.

\bibitem{agarwal2020flambe}
A.~Agarwal, S.~Kakade, A.~Krishnamurthy, and W.~Sun, ``Flambe: Structural complexity and representation learning of low rank mdps,'' in \emph{Advances in neural information processing systems}, Virtual Conference, December 2020, pp. 20\,095--20\,107.

\bibitem{9590452}
H.~Wu and X.~Luo, ``Instance-frequency-weighted regularized, nonnegative and adaptive latent factorization of tensors for dynamic qos analysis,'' in \emph{2021 IEEE International Conference on Web Services (ICWS)}, Chicago, IL, USA, 2021, pp. 560--568.

\bibitem{zeng2024novel}
N.~Zeng, X.~Li, P.~Wu, H.~Li, and X.~Luo, ``A novel tensor decomposition-based efficient detector for low-altitude aerial objects with knowledge distillation scheme,'' \emph{IEEE/CAA Journal of Automatica Sinica}, vol.~11, no.~2, pp. 487--501, 2024.

\bibitem{10159989}
Y.~Yuan, R.~Wang, G.~Yuan, and L.~Xin, ``An adaptive divergence-based non-negative latent factor model,'' \emph{IEEE Transactions on Systems, Man, and Cybernetics: Systems}, vol.~53, no.~10, pp. 6475--6487, 2023.

\bibitem{qin2023asynchronous}
W.~Qin and X.~Luo, ``Asynchronous parallel fuzzy stochastic gradient descent for high-dimensional incomplete data representation,'' \emph{IEEE Transactions on Fuzzy Systems}, vol.~32, no.~2, pp. 445--459, 2024.

\bibitem{10380219}
X.~Luo, J.~Chen, Y.~Yuan, and Z.~Wang, ``Pseudo gradient-adjusted particle swarm optimization for accurate adaptive latent factor analysis,'' \emph{IEEE Transactions on Systems, Man, and Cybernetics: Systems}, vol.~54, no.~4, pp. 2213--2226, 2024.

\bibitem{wu2023mmlf}
D.~Wu, P.~Zhang, Y.~He, and X.~Luo, ``Mmlf: Multi-metric latent feature analysis for high-dimensional and incomplete data,'' \emph{IEEE Transactions on Services Computing}, vol.~17, no.~2, pp. 575--588, 2024.

\bibitem{li2023generalized}
W.~Li, R.~Wang, and X.~Luo, ``A generalized nesterov-accelerated second-order latent factor model for high-dimensional and incomplete data,'' \emph{IEEE Transactions on Neural Networks and Learning Systems}, pp. 1--15, 2023.

\bibitem{luo2022neulft}
X.~Luo, H.~Wu, and Z.~Li, ``Neulft: A novel approach to nonlinear canonical polyadic decomposition on high-dimensional incomplete tensors,'' \emph{IEEE Transactions on Knowledge and Data Engineering}, vol.~35, no.~6, pp. 6148--6166, 2023.

\bibitem{wu2020advancing}
H.~Wu, X.~Luo, and M.~Zhou, ``Advancing non-negative latent factorization of tensors with diversified regularization schemes,'' \emph{IEEE Transactions on Services Computing}, vol.~15, no.~3, pp. 1334--1344, 2020.

\bibitem{luo2021adjusting}
X.~Luo, M.~Chen, H.~Wu, Z.~Liu, H.~Yuan, and M.~Zhou, ``Adjusting learning depth in nonnegative latent factorization of tensors for accurately modeling temporal patterns in dynamic qos data,'' \emph{IEEE Transactions on Automation Science and Engineering}, vol.~18, no.~4, pp. 2142--2155, 2021.

\bibitem{wu2020unified}
J.~Wu, X.~Xie, L.~Nie, Z.~Lin, and H.~Zha, ``Unified graph and low-rank tensor learning for multi-view clustering,'' in \emph{Proceedings of the AAAI conference on artificial intelligence}, Virtual Conference, February 2020, pp. 6388--6395.

\bibitem{fan2021lighter}
X.~Fan, Z.~Liu, J.~Lian, W.~X. Zhao, X.~Xie, and J.-R. Wen, ``Lighter and better: low-rank decomposed self-attention networks for next-item recommendation,'' in \emph{Proceedings of the 44th international ACM SIGIR conference on research and development in information retrieval}, Beijing, China, July 2021, pp. 1733--1737.

\bibitem{chen2020efficient}
C.~Chen, M.~Zhang, Y.~Zhang, Y.~Liu, and S.~Ma, ``Efficient neural matrix factorization without sampling for recommendation,'' \emph{ACM Transactions on Information Systems}, vol.~38, no.~2, pp. 1--28, 2020.

\bibitem{shi2022dual}
M.~Shi, X.~Liao, and Y.~Chen, ``A dual-population search differential evolution algorithm for functional distributed constraint optimization problems,'' \emph{Annals of Mathematics and Artificial Intelligence}, vol.~90, no.~10, pp. 1055--1078, 2022.

\bibitem{pilaszy2010fast}
I.~Pil{\'a}szy, D.~Zibriczky, and D.~Tikk, ``Fast als-based matrix factorization for explicit and implicit feedback datasets,'' in \emph{Proceedings of the fourth ACM conference on Recommender systems}, Valencia, Spain, September 2010, pp. 71--78.

\bibitem{shi2022particle}
M.~Shi, X.~Liao, and Y.~Chen, ``A particle swarm with local decision algorithm for functional distributed constraint optimization problems,'' \emph{International journal of pattern recognition and artificial intelligence}, vol.~36, no.~12, p. 2259025, 2022.

\bibitem{chen2017efficient}
J.~Chen, J.~Fang, W.~Liu, T.~Tang, X.~Chen, and C.~Yang, ``Efficient and portable als matrix factorization for recommender systems,'' in \emph{2017 IEEE International Parallel and Distributed Processing Symposium Workshops}, Orlando, Florida, USA, May 2017, pp. 409--418.

\bibitem{gemulla2011large}
R.~Gemulla, E.~Nijkamp, P.~J. Haas, and Y.~Sismanis, ``Large-scale matrix factorization with distributed stochastic gradient descent,'' in \emph{Proceedings of the 17th ACM SIGKDD international conference on Knowledge discovery and data mining}, San Diego, California, USA, August 2011, pp. 69--77.

\bibitem{2783258}
S.~Ahn, A.~Korattikara, N.~Liu, S.~Rajan, and M.~Welling, ``Large-scale distributed bayesian matrix factorization using stochastic gradient mcmc,'' in \emph{Proceedings of the 21th ACM SIGKDD International Conference on Knowledge Discovery and Data Mining}, ser. KDD '15.\hskip 1em plus 0.5em minus 0.4em\relax New York, NY, USA: Association for Computing Machinery, 2015, p. 9–18.

\bibitem{jin2016gpusgd}
J.~Jin, S.~Lai, S.~Hu, J.~Lin, and X.~Lin, ``Gpusgd: A gpu-accelerated stochastic gradient descent algorithm for matrix factorization,'' \emph{Concurrency and Computation: Practice and Experience}, vol.~28, no.~14, pp. 3844--3865, 2016.

\bibitem{li2017msgd}
H.~Li, K.~Li, J.~An, and K.~Li, ``Msgd: A novel matrix factorization approach for large-scale collaborative filtering recommender systems on gpus,'' \emph{IEEE Transactions on Parallel and Distributed Systems}, vol.~29, no.~7, pp. 1530--1544, 2017.

\bibitem{9001229}
S.~Zhou, R.~Kannan, and V.~K. Prasanna, ``Accelerating stochastic gradient descent based matrix factorization on fpga,'' \emph{IEEE Transactions on Parallel and Distributed Systems}, vol.~31, no.~8, pp. 1897--1911, 2020.

\bibitem{chin2015fast}
W.-S. Chin, Y.~Zhuang, Y.-C. Juan, and C.-J. Lin, ``A fast parallel stochastic gradient method for matrix factorization in shared memory systems,'' \emph{ACM Transactions on Intelligent Systems and Technology}, vol.~6, no.~1, pp. 1--24, 2015.

\bibitem{9664622}
H.~Wu, X.~Luo, M.~Zhou, M.~J. Rawa, K.~Sedraoui, and A.~Albeshri, ``A pid-incorporated latent factorization of tensors approach to dynamically weighted directed network analysis,'' \emph{IEEE/CAA Journal of Automatica Sinica}, vol.~9, no.~3, pp. 533--546, 2022.

\bibitem{10155460}
J.~Li, X.~Luo, Y.~Yuan, and S.~Gao, ``A nonlinear pid-incorporated adaptive stochastic gradient descent algorithm for latent factor analysis,'' \emph{IEEE Transactions on Automation Science and Engineering}, vol.~21, no.~3, pp. 3742--3756, 2024.

\bibitem{wu2023dynamic}
H.~Wu, X.~Wu, and X.~Luo, \emph{Dynamic Network Representation Based on Latent Factorization of Tensors}.\hskip 1em plus 0.5em minus 0.4em\relax Springer, 2023.

\bibitem{10580535}
Q.~Wang and H.~Wu, ``Dynamically weighted directed network link prediction using tensor ring decomposition,'' in \emph{2024 27th International Conference on Computer Supported Cooperative Work in Design (CSCWD)}, Tianjin, China, 2024, pp. 2864--2869.

\bibitem{recht2011hogwild}
B.~Recht, C.~Re, S.~Wright, and F.~Niu, ``Hogwild!: A lock-free approach to parallelizing stochastic gradient descent,'' in \emph{Advances in neural information processing systems}, Seattle, Washington, USA, December 2011.

\bibitem{zhuang2013fast}
Y.~Zhuang, W.-S. Chin, Y.-C. Juan, and C.-J. Lin, ``A fast parallel sgd for matrix factorization in shared memory systems,'' in \emph{Proceedings of the 7th ACM conference on Recommender systems}, Hong Kong, October 2013, pp. 249--256.

\bibitem{luo2012parallel}
X.~Luo, H.~Liu, G.~Gou, Y.~Xia, and Q.~Zhu, ``A parallel matrix factorization based recommender by alternating stochastic gradient decent,'' \emph{Engineering Applications of Artificial Intelligence}, vol.~25, no.~7, pp. 1403--1412, 2012.

\bibitem{9099403}
R.~Guo, F.~Zhang, L.~Wang, W.~Zhang, X.~Lei, R.~Ranjan, and A.~Y. Zomaya, ``Bapa: A novel approach of improving load balance in parallel matrix factorization for recommender systems,'' \emph{IEEE Transactions on Computers}, vol.~70, no.~5, pp. 789--802, 2021.

\bibitem{9727662}
H.~Wu, Y.~Xia, and X.~Luo, ``Proportional-integral-derivative-incorporated latent factorization of tensors for large-scale dynamic network analysis,'' in \emph{2021 China Automation Congress (CAC)}, Beijing, China, 2021, pp. 2980--2984.

\bibitem{qu2019accelerated}
G.~Qu and N.~Li, ``Accelerated distributed nesterov gradient descent,'' \emph{IEEE Transactions on Automatic Control}, vol.~65, no.~6, pp. 2566--2581, 2019.

\bibitem{8941240}
D.~Wu, Q.~He, X.~Luo, M.~Shang, Y.~He, and G.~Wang, ``A posterior-neighborhood-regularized latent factor model for highly accurate web service qos prediction,'' \emph{IEEE Transactions on Services Computing}, vol.~15, no.~2, pp. 793--805, 2022.

\bibitem{9159907}
D.~Wu, X.~Luo, M.~Shang, Y.~He, G.~Wang, and X.~Wu, ``A data-characteristic-aware latent factor model for web services qos prediction,'' \emph{IEEE Transactions on Knowledge and Data Engineering}, vol.~34, no.~6, pp. 2525--2538, 2022.

\bibitem{3472456}
Y.~Huang, Y.~Yin, Y.~Liu, S.~He, Y.~Bai, and R.~Li, ``A novel multi-cpu/gpu collaborative computing framework for sgd-based matrix factorization,'' in \emph{Proceedings of the 50th International Conference on Parallel Processing}, ser. ICPP '21.\hskip 1em plus 0.5em minus 0.4em\relax New York, NY, USA: Association for Computing Machinery, 2021.

\bibitem{nishioka2015scalable}
Y.~Nishioka and K.~Taura, ``Scalable task-parallel sgd on matrix factorization in multicore architectures,'' in \emph{2015 IEEE International Parallel and Distributed Processing Symposium Workshop}, Ann Arbor, MI, USA, May 2015, pp. 1178--1184.

\bibitem{oh2015fast}
J.~Oh, W.-S. Han, H.~Yu, and X.~Jiang, ``Fast and robust parallel sgd matrix factorization,'' in \emph{Proceedings of the 21th ACM SIGKDD International Conference on Knowledge Discovery and Data Mining}, Sydney, Australia, August 2015, pp. 865--874.

\bibitem{li2019mixture}
D.~Li, C.~Chen, T.~Lu, S.~M. Chu, and N.~Gu, ``Mixture matrix approximation for collaborative filtering,'' \emph{IEEE Transactions on Knowledge and Data Engineering}, vol.~33, no.~6, pp. 2640--2653, 2019.

\bibitem{ijcai2019-191}
G.~Guo, E.~Yang, L.~Shen, X.~Yang, and X.~He, ``Discrete trust-aware matrix factorization for fast recommendation,'' in \emph{Proceedings of the Twenty-Eighth International Joint Conference on Artificial Intelligence}, Macao, China, 7 2019, pp. 1380--1386.

\bibitem{9238448}
X.~Luo, Y.~Yuan, S.~Chen, N.~Zeng, and Z.~Wang, ``Position-transitional particle swarm optimization-incorporated latent factor analysis,'' \emph{IEEE Transactions on Knowledge and Data Engineering}, vol.~34, no.~8, pp. 3958--3970, 2022.

\bibitem{9152087}
Y.~Zhong, L.~Jin, M.~Shang, and X.~Luo, ``Momentum-incorporated symmetric non-negative latent factor models,'' \emph{IEEE Transactions on Big Data}, vol.~8, no.~4, pp. 1096--1106, 2022.

\bibitem{9072622}
Y.~Yuan, Q.~He, X.~Luo, and M.~Shang, ``A multilayered-and-randomized latent factor model for high-dimensional and sparse matrices,'' \emph{IEEE Transactions on Big Data}, vol.~8, no.~3, pp. 784--794, 2022.

\bibitem{9462337}
J.~Chen, X.~Luo, and M.~Zhou, ``Hierarchical particle swarm optimization-incorporated latent factor analysis for large-scale incomplete matrices,'' \emph{IEEE Transactions on Big Data}, vol.~8, no.~6, pp. 1524--1536, 2022.

\bibitem{9785520}
W.~Li, X.~Luo, H.~Yuan, and M.~Zhou, ``A momentum-accelerated hessian-vector-based latent factor analysis model,'' \emph{IEEE Transactions on Services Computing}, vol.~16, no.~2, pp. 830--844, 2023.

\bibitem{9357412}
M.~Shang, Y.~Yuan, X.~Luo, and M.~Zhou, ``An $\alpha$–$\beta$-divergence-generalized recommender for highly accurate predictions of missing user preferences,'' \emph{IEEE Transactions on Cybernetics}, vol.~52, no.~8, pp. 8006--8018, 2022.

\end{thebibliography}

\vspace{12pt}

\end{document}